\documentclass[runningheads]{llncs}

\let\vec\relax
\DeclareMathAccent{\vec}{\mathord}{letters}{"7E} 

\usepackage{graphicx}
\usepackage{comment}
\usepackage{amsmath,amssymb}
\usepackage{color}
\usepackage{lipsum}
\usepackage{tabularx}
\usepackage{bm}
\usepackage{multirow}
\usepackage{pifont}
\usepackage{capt-of}
\usepackage{hyperref}

\newcommand{\nicepar}[1]{\paragraph{\normalfont\textbf{#1}}}

\newcommand{\VKD}{\textbf{VKD}}

\renewcommand{\vec}[1]{\bm{#1}}

\newcommand{\mars}{MARS}
\newcommand{\duke}{Duke}
\newcommand{\dukefull}{Duke-Video-ReID}
\newcommand{\veri}{VeRi-776}
\newcommand{\atrw}{ATRW}

\newcommand{\argmin}{\operatornamewithlimits{argmin}}

\newcommand{\loss}[1]{\mathcal{L}_\textnormal{#1}}

\newcommand{\resnet}[1]{ResNet-#1}
\newcommand{\tea}[1]{\resnet{#1}}
\newcommand{\stf}[1]{ResVKD-#1}

\newcommand{\teadense}{DenseNet-121}
\newcommand{\stfdense}{DenseVKD-121}

\newcommand{\teamobile}{MobileNet-V2}
\newcommand{\stfmobile}{MobileVKD-V2}
\newcommand{\R}{\mathbb{R}}

\newcommand{\topk}[1]{top$_{#1}$}

\newcommand{\cmark}{\ding{51}}%
\newcommand{\xmark}{\ding{55}}%

\newcommand{\attnp}[1]{\includegraphics[width=0.08\columnwidth]{img/attn_grid/#1.png}}
\newcommand{\attnv}[1]{\includegraphics[width=0.1\columnwidth]{img/attn_grid/#1.png}}
\newcommand{\attnt}[1]{\includegraphics[width=0.2\columnwidth]{img/attn_grid/#1.png}}

\newcommand{\trackpersonimg}[1]{\includegraphics[width=0.11\columnwidth]{img/dist_matrix_person/#1.png}}
\newcommand{\trackvehicleimg}[1]{\includegraphics[width=0.156\columnwidth]{img/dist_matrix_vehicle/#1.png}}

\newcommand{\reid}{Re-ID}
\newcommand{\reidlong}{Re-Identification}

\newcommand{\quotationmarks}[1]{``#1''}

\begin{document}

\pagestyle{headings}
\mainmatter
\def\ECCVSubNumber{996}  

\title{Robust Re-Identification by Multiple Views Knowledge Distillation}

\titlerunning{Robust Re-Identification by Multiple Views Knowledge Distillation}
\author{Angelo Porrello \orcidID{0000-0002-9022-8484},
Luca Bergamini \orcidID{0000-0003-1221-8640},
Simone Calderara \orcidID{0000-0001-9056-1538}}
\authorrunning{A. Porrello, L. Bergamini, S. Calderara}
\institute{AImageLab, University of Modena and Reggio Emilia\\
\email{\{angelo.porrello, luca.bergamini24, simone.calderara\}@unimore.it}
}

\maketitle
\begin{abstract}
To achieve robustness in Re-Identification, standard methods leverage tracking information in a Video-To-Video fashion. However, these solutions face a large drop in performance for single image queries (e.g., Image-To-Video setting). Recent works address this severe degradation by transferring \textit{temporal information} from a Video-based network to an Image-based one. In this work, we devise a training strategy that allows the transfer of a superior knowledge, arising from a set of views depicting the target object. Our proposal -- Views Knowledge Distillation (VKD) -- pins this \textit{visual variety} as a supervision signal within a teacher-student framework, where the teacher educates a student who observes fewer views. As a result, the student outperforms not only its teacher but also the current state-of-the-art in Image-To-Video by a wide margin (6.3\% mAP on \mars{}, 8.6\% on \duke{} and 5\% on \veri{}). A thorough analysis -- on Person, Vehicle and Animal Re-ID -- investigates the properties of VKD from a qualitatively and quantitatively perspective. Code is available at \href{https://github.com/aimagelab/VKD}{https://github.com/aimagelab/VKD}.
\end{abstract}
\section{Introduction}
Recent advances on Metric Learning~\cite{schroff2015metric1-facenet,sohn2016metric2-npair,wang2017metric3-angular,ustinova2016metric4-hist} give to researchers the foundation for computing suitable distance metrics between data points. In this context, \reidlong{} (Re-ID) has greatly benefited in diverse domains~\cite{zheng2016survey-person,khan2019survey-vehicle,schneider2019survey-animal}, as the common paradigm requires distance measures exhibiting robustness to variations in background clutters, as well as different viewpoints. To meet these criteria, various deep learning based approaches leverage videos to provide detailed descriptions for both query and gallery items. However, such a setting -- known as Video-To-Video (V2V) \reid{} -- does not represent a viable option in many scenarios (e.g. surveillance)~\cite{zhang2017lstm,xie2019i2vreason1,nguyen2018i2vreason2,gu2019TKP}, where the query comprises a single image (Image-To-Video, I2V). 

\begin{figure}[t]
    \begin{minipage}[t][][b]{.50\linewidth}
    \centering
    \vskip 0.15in
    \begin{tabular}{llllllll}
         &  \trackpersonimg{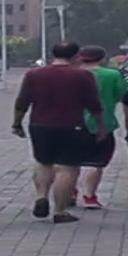} & \trackpersonimg{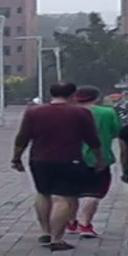} &
        \trackpersonimg{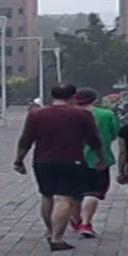} &  \trackpersonimg{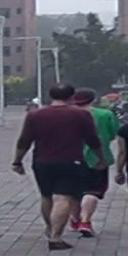} & \trackpersonimg{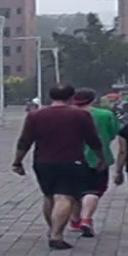} &  \trackpersonimg{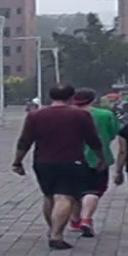} & \trackpersonimg{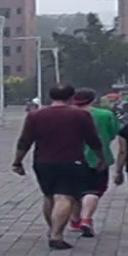}\\
        \hline
        \noalign{\vskip 0.05in}  
        \end{tabular}
        \begin{tabular}{llllll} 
         &  \trackvehicleimg{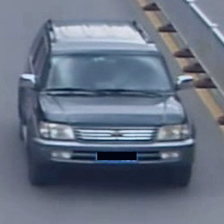} & \trackvehicleimg{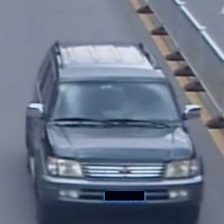} &  \trackvehicleimg{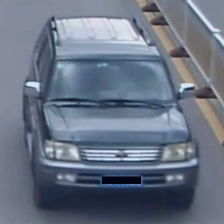} & \trackvehicleimg{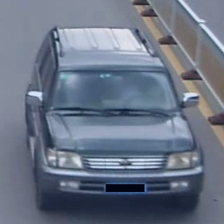} &  \trackvehicleimg{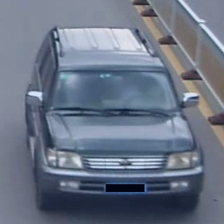} \\
    \end{tabular}
    \\\small{(a) Two examples of tracklets}.
    \end{minipage} 
    \hfill
     \begin{minipage}[t][][b]{.48\linewidth}
    \centering
    \setlength{\tabcolsep}{3pt}
            \begin{tabular}{c|c}
            \textbf{\mars{}} & \textbf{\veri{}} \\
            \includegraphics{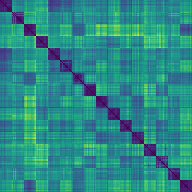} & \includegraphics{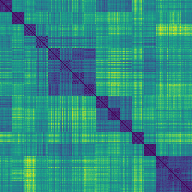} \\
            \end{tabular}
            \\\small{(b) Distances between tracklets features.}
    \end{minipage}
    \vskip.5\baselineskip%
    \hfill
    \rule{345pt}{0.5pt}
    \\
    \begin{minipage}[t][][b]{.50\linewidth}
    \centering
    \begin{tabular}{llllllll}
 &  \trackpersonimg{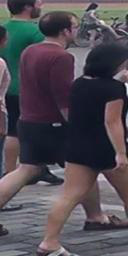} & \trackpersonimg{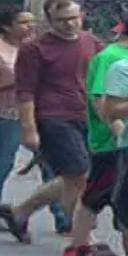} &
\trackpersonimg{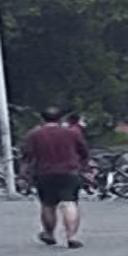} &  \trackpersonimg{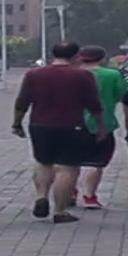} & \trackpersonimg{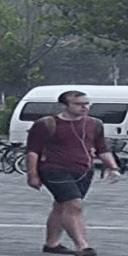} &  \trackpersonimg{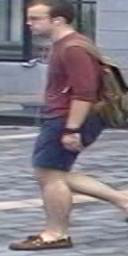} & \trackpersonimg{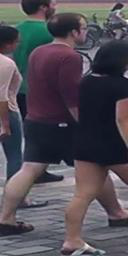}\\
\hline
\noalign{\vskip 0.05in} 
\end{tabular}
\begin{tabular}{llllll}
 & \trackvehicleimg{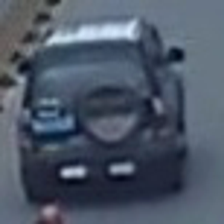} & \trackvehicleimg{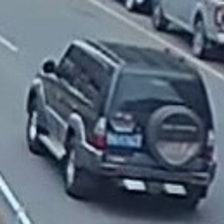} &  \trackvehicleimg{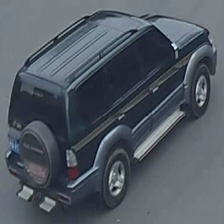} & \trackvehicleimg{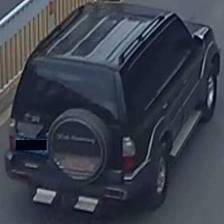}  & \trackvehicleimg{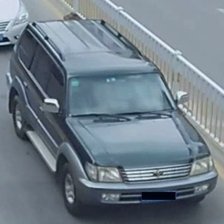}\\
\end{tabular}
    \\\small{(c) Two examples of multiview sets.}
    \end{minipage} 
    \hfill
     \begin{minipage}[t][][b]{.48\linewidth}
    \centering
    \setlength{\tabcolsep}{3pt}
            \begin{tabular}{c|c}
            \includegraphics{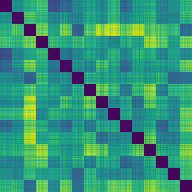} &
            \includegraphics{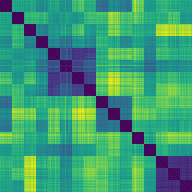}
            \end{tabular}
    \\\small{(d) Distances when ensambling views.}
    \label{fig:time_vs_mv}
    \end{minipage}
    \caption{Visual comparison between tracklets and viewpoints variety, on person (\mars{}~\cite{zheng2016mars}) and vehicle (\veri{}~\cite{liu2016veri}) re-id. Right: pairwise distances computed on top of features from \resnet{50}. Inputs batches comprise 192 sets from 16 different identities, grouped by ground truth identity along each axis.}
\hfill
\end{figure}

\noindent As observed in~\cite{gu2019TKP}, a large gap in Re-ID performance still subsists between V2V and I2V, highlighting the number of query images as a critical factor in achieving good results. Contrarily,
we advise the learnt representation should not be heavily affected when few images are shown to the network (\textit{e.g.} only one). To bridge such a gap,~\cite{gu2019TKP,bhardwaj2019fewerframes} propose a teacher-student paradigm, in which the student -- in contrast with the teacher -- has access to a small fraction of the frames in the video. Since the student is educated to mimic the output space of its teacher, it will show higher generalisation properties than its teacher when a single frame is available. It is noted that these approaches rely on transferring \textit{temporal} information: as datasets often come with tracking annotation, they can guide the transfer from a tracklet into one of its frames. In this respect, we argue the limits of transferring temporal information: in fact, it is reasonable to assume an high correlation between frames from the same tracklet (Fig.~\ref{fig:time_vs_mv}a), which may potentially underexploit the transfer. Moreover, limiting the analysis to the temporal domain does not guarantee robustness to variation in background appearances.

\noindent Here, we make a step forward and consider which information to transfer, shifting the paradigm from \textit{time} to \textit{views}: we argue that more valuable information arises when ensembling diverse views of the same target (Fig.~\ref{fig:time_vs_mv}c). This information often comes for free, as various datasets~\cite{zheng2016mars,wu2018duke1,liu2016veri,bergamini2018multi} provide images capturing the same target from different camera viewpoints. To support our claim, Fig. \ref{fig:time_vs_mv} (right) reports pairwise distances computed on top of \resnet{50}, when trained on Person and Vehicle Re-ID. In more details: matrices from Fig.~\ref{fig:time_vs_mv}b visualise the distances when tracklets are provided as input, whereas Fig.~\ref{fig:time_vs_mv}d shows the same for sets of views. As one can see, leveraging different views leads to a more distinctive blockwise pattern: namely, activations from the same identity are more consistent if compared to the ones computed in the tracklet scenario. As shown in~\cite{tung2019similarity}, this reflects a higher capacity to capture the semantics of the dataset, and therefore a \textit{graceful} knowledge a teacher can transfer to a student.

\noindent Based on the above, we propose Views Knowledge Distillation (\VKD{}), which transfers the knowledge lying in several views in a teacher-student fashion. VKD devises a two-stage procedure, which pins the visual variety as a teaching signal for a student who has to recover it using fewer views. We remark the following contributions: \textit{i)} the student outperforms its teacher by a large margin, especially in the Image-To-Video setting; \textit{ii)} a thorough investigation shows that the student focuses more on the target compared to its teacher and discards uninformative details; \textit{iii)} importantly, we do not limit our analysis to a single domain, but instead achieve strong results on Person, Vehicle and Animal Re-ID.
\section{Related Works}
\nicepar{Image-To-Video \reidlong{}} The I2V \reid{} task has been successfully applied to multiple domains. In person \reid{},~\cite{wang2017p2snet} frames it as a point-to-set task, where image and video domains are aligned using a single deep network. The authors of~\cite{zhang2017lstm} exploit time information by aggregating frames features via a Long-Short Term Memory. Eventually, a dedicated sub-network aggregates video features and match them against single image query ones. Authors of MGAT~\cite{bao2019maskedGAT} employ a Graph Neural Network to model relationships between samples from different identities, thus enforcing similarity in the feature space. Dealing with vehicle \reid{}, authors from~\cite{liu2017provid} introduce a large-scale dataset (\veri{}) and propose PROVID and PROVID-BOT, which combine appearance and plate information in a progressive fashion. Differently, RAM~\cite{liu2018ram} exploits multiple branches to extract global and local features, imposing a separate supervision on each branch and devising an additional one to predict vehicle attributes. VAMI~\cite{zhou2018vami} employs a viewpoint aware attention model to select core regions for different viewpoints. At inference time, they obtain a multiview descriptor through a conditional generative network, inferring information regarding the unobserved viewpoints. Differently, our approach asks the student to do it implicitly and in a lightweight fashion, thus avoiding the need for additional modules. Similarly to VAMI,~\cite{chu2019vanet} predicts the vehicle viewpoint along with appearance features; at inference, the framework provides distances according to the predicted viewpoint.
\nicepar{Knowledge Distillation} has been first investigated in~\cite{romero2014fitnets,hinton2015distilling,zagoruyko2016payingattention} for model compression: the idea is to instruct a lightweight model (student) to mimic the capabilities of a deeper one (teacher): as a gift, one could achieve both an acceleration in inference time as well as a reduction in memory consumption, without experiencing a large drop in performance. In this work, we benefit from the techniques proposed in~\cite{hinton2015distilling,tung2019similarity} for a different purpose: we are not primarily engaged in educating a lightweight module, but on improving the original model itself. In this framework -- often called \textit{self-distillation}~\cite{furlanello2018born,yang2018tolerantteacher} -- the transfer occurs from the teacher to a student with the same architecture, with the aim of improving the overall performance at the end of the training. Here, we get a step ahead and introduce an asymmetry between the teacher and student, which has access to fewer frames. In this respect, our work closely relates to what~\cite{bhardwaj2019fewerframes} devises for Video Classification. Besides facing another task, a key difference subsists: while~\cite{bhardwaj2019fewerframes} limits the transfer along the temporal axis, our proposal advocates for distilling many views into fewer ones. On this latter point, we shall show that the teaching signal can be further enhanced when opening to diverse camera viewpoints. In the \reidlong{} field, Temporal Knowledge Propagation (TKP)~\cite{gu2019TKP} similarly exploits intra-tracklet information to encourage the image-level representations to approach the video-level ones. In contrast with TKP: \textit{i)} we do not rely on matching internal representations but instead their distances solely, thus making our proposal viable for cross-architecture transfer too; \textit{ii)} at inference time, we make use of a single shared network to deal with both image and video domains, thus halving the number of parameters; \textit{iii)} during transfer, we benefit from a larger visual variety, emerging from several viewpoints.
\section{Method}
\begin{figure}[t]
    \centering
    \includegraphics[width=\textwidth]{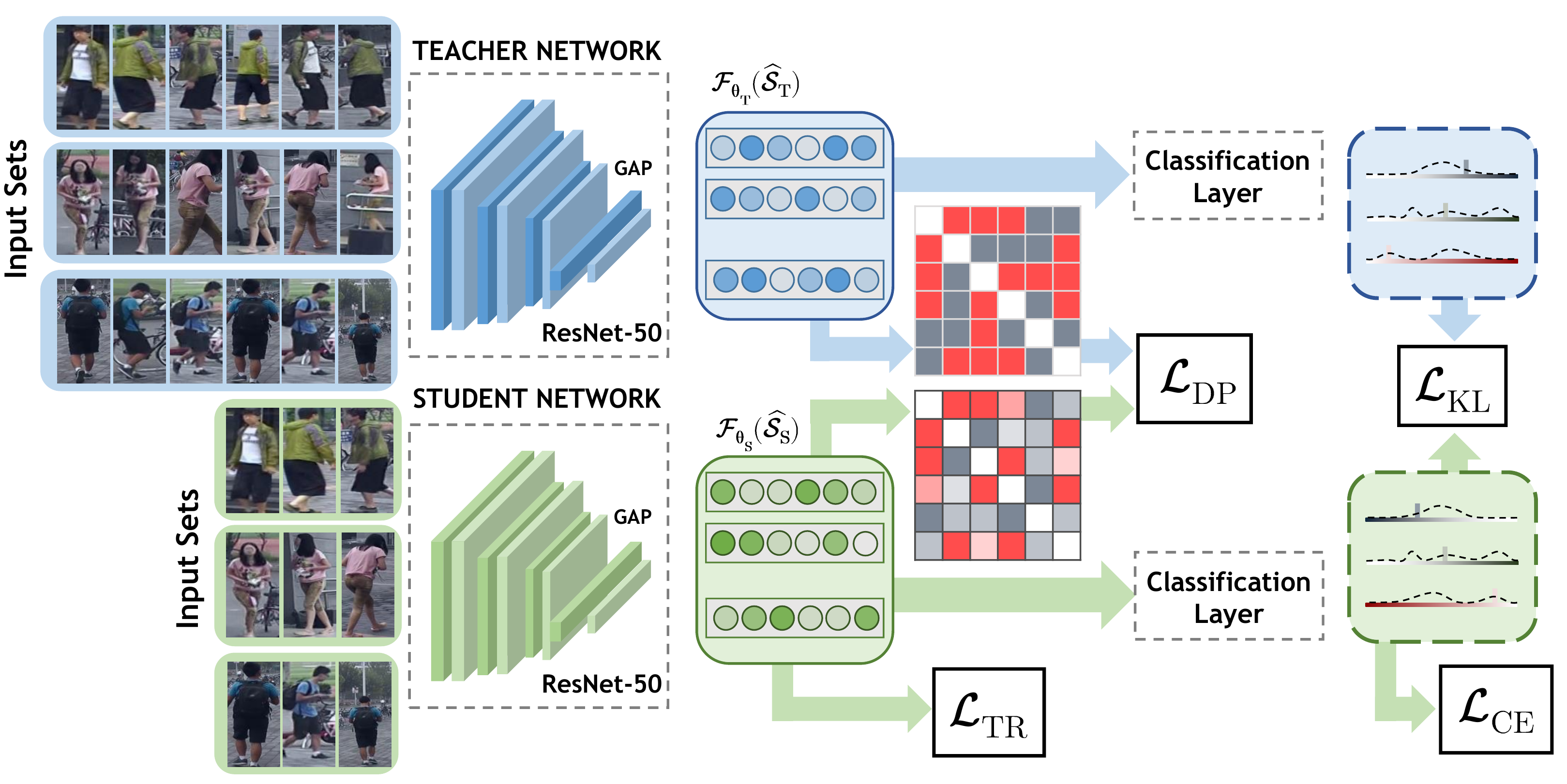}
    \caption{An overview of Views Knowledge Distillation (VKD): a student network is optimised to mimic the behaviour of its teacher using fewer views.}
    \label{fig:framework}
\end{figure}
\noindent We purse the aim of learning a function $\mathcal{F}_{\theta}(\mathcal{S})$ mapping a set of images $\mathcal{S} = (s_1, s_2, ..., s_n)$ into a representative embedding space. Specifically, $\mathcal{S}$ is a sequence of bounding boxes crops depicting a target (\textit{e.g.} a person or a car), for which we are interested in inferring its corresponding identity. We take advantage of Convolutional Neural Networks (CNNs) for modelling $\mathcal{F}_{\theta}(\mathcal{S})$. Here, we look for two distinctive properties, aspiring to representations that are \textit{i)} invariant to differences in background and viewpoint and \textit{ii)} robust to a reduction in the number of query images. To achieve this, our proposal frames the training algorithm as a two-stage procedure, as follows: 
\begin{itemize}
    \item \textbf{First step} (Sec.~\ref{subsec:teacher}): the backbone network is trained for the standard Video-To-Video setting. 
    \item \textbf{Second step} (Sec.~\ref{sub:distillation}): we appoint it as the teacher and freeze its parameters. Then, a new network with the role of the student is instantiated. As depicted in Fig.~\ref{fig:framework}, we feed frames representing different views as input to the teacher and ask the student to mimic the same outputs from fewer frames.
\end{itemize}
\subsection{Teacher Network}
\label{subsec:teacher}
\noindent Without loss of generality, we will refer to \resnet{50}~\cite{he2016deep} as the backbone network, namely a module $f_{\theta}: \R^{\text{W} \times \text{H} \times 3} \mapsto \R^{\text{D}}$ mapping each image $s_{i}$ from $S$ to a fixed-size representation $d_{i}$ (in this case $D = 2048$). Following previous works~\cite{luo2019bag,gu2019TKP}, we initialise the network weights on ImageNet and additionally include few amendments~\cite{luo2019bag} to the architecture. First, we discard both the last ReLU activation function and final classification layer in favour of the BNNeck one~\cite{luo2019bag} (\textit{i.e.} batch normalisation followed by a linear layer). Second: to benefit from fine-grained spatial details, the stride of the last residual block is decreased from 2 to 1.
%
%
\nicepar{Set representation} Given a set of images $S$, several solutions~\cite{liu2017quality,zhang2017lstm,liu2019NVAN} may be assessed for designing the aggregation module, which fuses a variable-length set of representations $d_1, d_2, \dots, d_n$ into a single one. Here, we naively compute the set-level embedding $\mathcal{F}(\mathcal{S})$ through a temporal average pooling. While we acknowledge better aggregation modules exist, we do not place our focus on devising a new one, but instead on improving the earlier features extractor.
\nicepar{Teacher optimisation} We train the base network - which will be the teacher during the following stage - combining a classification term $\loss{CE}$ (cross-entropy) with the triplet loss $\loss{TR}$\footnote{For the sake of clarity, all the loss terms are referred to one single example. In the implementation, we extend the penalties to a batch by averaging.}. The first can be formulated as:
\begin{equation}\label{eq:loss_ce}
    \loss{CE} = -\vec{y} \log {\vec{\hat{y}}}
\end{equation}
\noindent where $\vec{y}$ and $\vec{\hat{y}}$ represent the one-hot labels (identities) and the output of the softmax respectively. The second term $\loss{TR}$ encourages distance constraints in feature space, moving closer representations from the same target and pulling away ones from different targets. Formally:
\begin{equation}\label{eq:loss_tr}
    \loss{TR} = \ln(1 + e^{\mathcal{D}\left(\mathcal{F}_{\theta}(\mathcal{S}^i_a), \mathcal{F}_{\theta}(\mathcal{S}^i_p)\right) - \mathcal{D}\left(\mathcal{F}_{\theta}(\mathcal{S}^i_a), \mathcal{F}_{\theta}(\mathcal{S}^j_n)\right)}), 
\end{equation}
\noindent where $\mathcal{S}_p$ and $\mathcal{S}_n$ are the hardest positive and negative for an anchor $\mathcal{S}_a$ within the batch. In doing so, we rely on the batch hard strategy~\cite{hermans2017defensetriplet} and include P identities coupled with K samples in each batch. Importantly, each set $\mathcal{S}^i$ comprises images drawn from the same tracklet~\cite{liu2019NVAN,fu2019STA}.
\subsection{Views Knowledge Distillation (\VKD{})}\label{sub:distillation}
After training the teacher, we propose to enrich its representation capabilities, especially when only few images are made available to the model. To achieve this, our proposal bets on the knowledge we can gather from different views, depicting the same object under different conditions. When facing re-identification tasks, one can often exploit camera viewpoints~\cite{zheng2016mars,ristani2016duke,liu2016veri} to provide a larger variety of appearances for the target identity. Ideally, we would like to teach a new network to recover such a variety even from a single image. Since this information may not be inferred from a single frame, this can lead to an ill-posed task. Still, one can underpin this knowledge as a supervision signal, encouraging the student to focus on important details and favourably discover new ones. On this latter point, we refer the reader to Section~\ref{sub:vkdablations} for a comprehensive discussion.

\noindent Views Knowledge Distillation (\VKD{}) stresses this idea by forcing a student network $\mathcal{F}_{\theta_{S}}(\cdot)$ to match the outputs of the teacher $\mathcal{F}_{\theta_{T}}(\cdot)$. In doing so, we: \textit{i)} allow the teacher to access frames $\hat{\mathcal{S}_{T}} = (\hat{s}_1, \hat{s}_2, \dots, \hat{s}_{N})$ from different viewpoints; \textit{ii)} force the student to mimic the teacher output starting from a subset $\hat{\mathcal{S}_{S}} = (\hat{s}_1, \hat{s}_2, \dots, \hat{s}_{M}) \subset \hat{\mathcal{S}_{T}}$ with cardinality $M < N$ (in our experiments, $M = 2$ and $N = 8$). The frames in $\hat{\mathcal{S}_{S}}$ are uniformly sampled from $\hat{\mathcal{S}_{T}}$ without replacement. This asymmetry between the teacher and the student leads to a self-distillation objective, where the latter can achieve better solutions despite inheriting the same architecture of the former. 

\noindent To accomplish this, VKD exploits the Knowledge Distillation loss~\cite{hinton2015distilling}:
\begin{equation}\label{eq:loss_kd}
    \loss{KD} = \tau^2 \ \text{KL}(\vec{y}_{T}\parallel\vec{y}_{S}) 
\end{equation}
\noindent where $\vec{y}_{T}=\text{softmax}(\vec{h}_{T} / \tau)$ and $\vec{y}_{S}=\text{softmax}(\vec{h}_{S} / \tau)$ are the distributions -- smoothed by a temperature $\tau$ -- we attempt to match\footnote{Since the teacher parameters are fixed, its entropy is constant and the objective of Eq.~\ref{eq:loss_kd} reduces to the cross-entropy between $\vec{y}_{T}$ and $\vec{y}_{S}$.}. Since the student experiences a different task from the teacher one, Eq.~\ref{eq:loss_kd} resembles the regularisation term imposed by~\cite{li2016learning} to relieve \textit{catastrophic forgetting}. In a similar vein, we intend to \textit{strengthen} the model in the presence of few images, whilst not \textit{deteriorating} the capabilities it achieved with longer sequences.

\noindent In addition to fitting the output distribution of the teacher (Eq.~\ref{eq:loss_kd}), our proposal devises additional constraints on the embedding space learnt by the student. In details, VKD encourages the student to mirror the pairwise distances spanned by the teacher. Indicating with $\mathcal{D}_{T}[i,j] \equiv \mathcal{D}(\mathcal{F}_{\theta_{T}}(\hat{\mathcal{S}_{T}}[i]),\mathcal{F}_{\theta_{T}}(\hat{\mathcal{S}_{T}}[j]))$ the distance induced by the teacher between the $i$-th and $j$-th sets (the same notation $\mathcal{D}_{S}[i,j]$ also holds for the student), VKD seeks to minimise: 
\begin{equation}\label{eq:loss_dp}
    \loss{DP} = \sum _{(i,j) \in\,\binom{B}{2}} (\mathcal{D}_{T}[i,j] - \mathcal{D}_{S}[i,j])^2,
\end{equation}
\noindent where $B$ equals the batch size. Since the teacher has access to several viewpoints, we argue that distances spanned in its space yield a powerful description of corresponding identities. From the student perspective, distances preservation provides additional semantic knowledge. Therefore, this holds an effective supervision signal, whose optimisation is made more challenging since fewer images are available to the student. 

\noindent Even thought VKD focuses on \textit{self-distillation}, we highlight that both $\loss{KD}$ and $\loss{DP}$ allow to match models with different embedding size, which would not be viable under the minimisation performed by~\cite{gu2019TKP}. As an example, it is still possible to distill \resnet{101} ($D = 2048$) into \teamobile{}~\cite{sandler2018mobilenetv2} ($D = 1280$).

\nicepar{Student optimisation} The VKD overall objective combines the distillation terms ($\loss{KD}$ and $\loss{DP}$) with the ones optimised by the teacher - $\loss{CE}$ and $\loss{TR}$ - that promote higher conditional likelihood w.r.t.\ ground truth labels. To sum up, VKD aims at strengthening the features of a CNN in Re-ID settings through the following optimisation problem:
\begin{equation}\label{eq:loss_total}
    \argmin_{\theta_S}\quad\loss{VKD} \equiv \loss{CE} + \loss{TR} + \alpha \loss{KD} + \beta \loss{DP}, 
\end{equation}
\noindent where $\alpha$ and $\beta$ are two hyperparameters balancing the contributions to the total loss $\loss{VKD}$. We conclude with a final note on the student initialisation: we empirically found beneficial to start from the teacher weights $\theta_T$ except for the last convolutional block, which is reinitialised according to the ImageNet pretraining. We argue this represents a good compromise between exploring new configurations and exploiting the abilities already achieved by the teacher.
\section{Experiments}
\label{sec:experiments}
\nicepar{Evaluation Protocols} We indicate the query-gallery matching as x2x, where both x terms are features that can be generated by either a single (I) or multiple frames (V). In the \textbf{Image-to-Image (I2I)} setting features extracted from a query set image are matched against features from individual images in the gallery. This protocol -- which has been amply employed for person \reid{} and face recognition -- has a light impact in terms of resources footprint. However, a single image captures only a single view of the identity, which may not be enough for identities exhibiting multi-modal distributions. Contrarily, the \textbf{Video-to-Video (V2V)} setting enables to capture and combine different modes in the input, but with a significant increase in the number of operations and memory. Finally, the \textbf{Image-to-Video (I2V)} setting~\cite{zhou2018SCCN,zhou2018vami,liu2018ram,wang2017oife,liu2017provid} represents a good compromise: building the gallery may be slow, but it is often performed offline. Moreover, matchings perform extremely fast, as a query comprise only a single image. We remark that \textit{i)} We adopt the standard \quotationmarks{\textit{Cross Camera Validation}} protocol, not considering examples of the gallery from the same camera of the query at evaluation and \textit{ii)} even if VKD relies on frames from different camera during train, we strictly adhere to the common schema and switch to tracklet-based inputs at evaluation time.
\nicepar{Evaluation Metrics} While settings vary between different dataset, evaluation metrics for \reidlong{} are shared by the vast majority of works in the field. In the followings, we report performance in terms of top-k accuracy and Mean Average Precision (mAP). By combining them, we evaluate VKD both in terms of accuracy and ranking performance. 
\subsection{Datasets\label{sub:datasets}}
\noindent \textbf{Person Re-ID}: \textbf{\mars{}}~\cite{zheng2016mars} comprises 19680 tracklets from 6 different cameras, capturing 1260 different identities (split between 625 for the training set, 626 for the gallery and 622 for the query) with 59 frames per tracklet on average. MARS has shown to be a challenging dataset because it has been automatically annotated, leading to errors and false detections~\cite{zheng2016survey-person}. The \textbf{\duke{}}~\cite{ristani2016duke} dataset was first introduced for multi-target and multi-camera surveillance purposes, and then expanded to include person attributes and identities (414 ones). Consistently with~\cite{gu2019TKP,si2018DuATM,liu2019NVAN,matiyali2019clip-simil}, we use the \textbf{\dukefull{}}~\cite{wu2018duke1} variant, where identities have been manually annotated from tracking information\footnote{In the following, we refer to \dukefull{} simply as \duke{}. Another variant of \duke{} named Duke-ReID exists~\cite{ristani2018duke2}, but it does not come with query tracklets.}. It comprises 5534 video tracklets from 8 different cameras, with 167 frames per tracklet on average. Following~\cite{gu2019TKP}, we extract the first frame of every tracklet when testing in the I2V setting, for both \mars{} and \duke{}.

\noindent \textbf{Vehicle Re-ID}: \textbf{\veri{}}~\cite{liu2016veri} has been collected from 20 fixed cameras, capturing vehicles moving on a circular road in a $1.0$ $\text{km}^2$ area. It contains 18397 tracklets with an average number of 6 frames per tracklet, capturing 775 identities split between train (575) and gallery (200). The query set shares identities consistently with the gallery, but differently from the other two sets it includes only a single image for each couple (id, camera). Consequently, all recent methods perform the evaluation following the I2V setting.

\noindent \textbf{Animal Re-ID}: The \textbf{Amur Tiger}~\cite{li2019atrw} \reidlong{} in the Wild (\atrw{}) is a recently introduced dataset collected from a diverse set of wild zoos. The training set includes 107 subjects and 17.6 images on average per identity; no information is provided to aggregate images into tracklets. It is possible to evaluate only the I2I setting through a remote http server. As done in~\cite{liu2019tiger-top1}, we horizontally flip the training images to duplicate the number of identities available, thus resulting in 214 training identities. 

\nicepar{Implementation details} Following~\cite{hermans2017defensetriplet,liu2019NVAN} we adopt the following hyperparameters for \mars{} and \duke{}: \textit{i)} each batch contains $P=8$ identities with $K=4$ samples each; \textit{ii)} each sample comprises 8 images equally spaced in a tracklet. Differently, for image-based datasets (\atrw{} and \veri{}) we increase $P$ to $18$ and use a single image at a time. All the teacher networks are trained for 300 epoch using Adam~\cite{kingma2014adam}, setting the learning rate to $10^{-4}$ and multiplying it by $0.1$ every 100 epochs. During the distillation stage, we feed $N=8$ images to the teacher and $M=2$ ones (picked at random) to the student. We found beneficial to train the student longer: so, we set the number of epochs to 500 and the learning rate decay steps at 300 and 450. We keep fixed $\tau=10$ (Eq.~\ref{eq:loss_kd}), $\alpha=10^{-1}$ and $\beta=10^{-4}$ (Eq.~\ref{eq:loss_total}) in all experiments. To improve generalisation, we apply data augmentation as described in~\cite{luo2019bag}. Finally, we put the teacher in training mode during distillation (consequently, batch normalisation~\cite{ioffe2015batch} statistics are computed on a batch basis): as observed in~\cite{bagherinezhad2018label}, this provides more accurate teacher labels.
\subsection{Self-Distillation}
\begin{figure}[t]
    \centering
    \includegraphics[width=1.0\columnwidth]{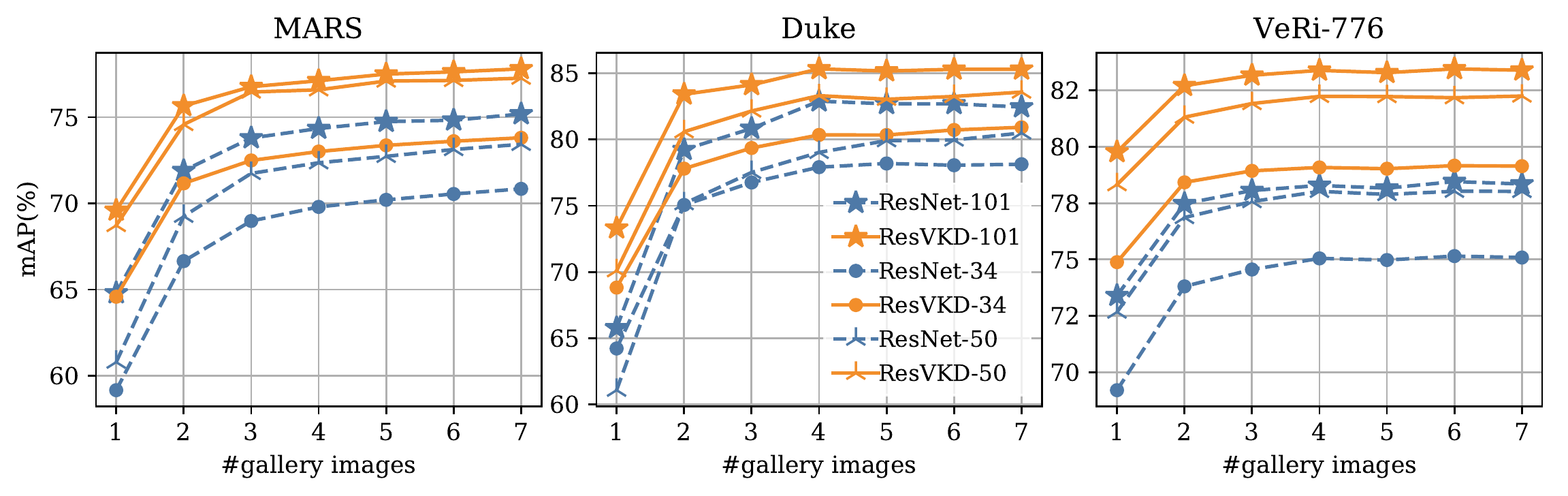}
    \caption{Performance (mAP) in the Image-To-Video setting when changing at evaluation time the number of frames in each gallery tracklet.}
    \label{fig:plot_ngallery}
\end{figure}
\begin{table}[t]
\caption{Self-Distillation results across datasets, settings and architectures.}\label{tab:self_distill}
\centering
\setlength{\tabcolsep}{2pt}
\resizebox{0.98\textwidth}{!}{
\begin{tabular}{l|cc|cc|cc|cc|cc|cc}
    \hline
    & \multicolumn{4}{c|}{\mars{}} & \multicolumn{4}{c|}{\duke{}} & \multicolumn{4}{c}{\veri{}}\\\cline{2-13}
    & \multicolumn{2}{c|}{I2V} & \multicolumn{2}{c|}{V2V}  & \multicolumn{2}{c|}{I2V} & \multicolumn{2}{c|}{V2V} & \multicolumn{2}{c|}{I2I} & \multicolumn{2}{c}{I2V}\\
    & cmc1 & mAP & cmc1 & mAP & cmc1 & mAP & cmc1 & mAP & cmc1 & mAP & cmc1 & mAP\\
    \hline
    \tea{34} & 80.81 & 70.74 & 86.67 & 78.03 & 81.34 & 78.70 & 93.45 & \textbf{91.88} & 92.97 & 70.30 & 93.80 & 75.01\\
    \stf{34} & \textbf{82.17} & \textbf{73.68} & \textbf{87.83} & \textbf{79.50} & \textbf{83.33} & \textbf{80.60} & \textbf{93.73} & 91.62 & \textbf{95.29} & \textbf{75.97} & \textbf{94.76} & \textbf{79.02}\\
    \hline
    \tea{50} & 82.22 & 73.38 & 87.88 & 81.13 & 82.34 & 80.19 & \textbf{95.01} & \textbf{94.17} & 93.50 & 73.19 & 93.33 & 77.88\\
    \stf{50} & \textbf{83.89} & \textbf{77.27} & \textbf{88.74} & \textbf{82.22} & \textbf{85.61} & \textbf{83.81} & \textbf{95.01} & 93.41 & \textbf{95.23} & \textbf{79.17} & \textbf{95.17} & \textbf{82.16}\\
    \hline
    \tea{101} & 82.78 & 74.94 & 88.59 & 81.66 & 83.76 & 82.89 & \textbf{96.01} & \textbf{94.73} & 94.28 & 74.27 & 94.46 & 78.20\\
    \stf{101} & \textbf{85.91} & \textbf{77.64} & \textbf{89.60} & \textbf{82.65} & \textbf{86.32} & \textbf{85.11} & 95.44 & 93.67 & \textbf{95.53} & \textbf{80.62} & \textbf{96.07} & \textbf{83.26}\\
    \hline
    \hline
    \tea{50bam} & 82.58 & 74.11 & 88.54 & 81.19 & 82.48 & 80.24 & 94.87 & \textbf{93.82} & 93.33 & 72.73 & 93.80 & 77.14\\
    \stf{50bam} & \textbf{84.34} & \textbf{78.13} & \textbf{89.39} & \textbf{83.07} & \textbf{86.18} & \textbf{84.54} & \textbf{95.16} & 93.45 & \textbf{96.01} & \textbf{78.67} & \textbf{95.71} & \textbf{81.57}\\
    \hline
    \hline
    \teadense{} & 82.68 & 74.34 & 89.75 & 81.93 & 82.91 & 80.26 & 93.73 & 91.73 & 91.24 & 69.24 & 91.84 & 74.52 \\
    \stfdense{} & \textbf{84.04} & \textbf{77.09} & \textbf{89.80} & \textbf{82.84} & \textbf{86.47} & \textbf{84.14} & \textbf{95.44} & \textbf{93.54} & \textbf{94.34} & \textbf{76.23} & \textbf{93.80} & \textbf{79.76} \\
    \hline
    \hline
    \teamobile{}{} & 78.64 & 67.94 & 85.96 & 77.10 & 78.06 & 74.73 & 93.30 & 91.56 & 88.80 & 64.68 & 89.81 & 69.90\\
    \stfmobile{} & \textbf{83.33} & \textbf{73.95} & \textbf{88.13} & \textbf{79.62} & \textbf{83.76} & \textbf{80.83} & \textbf{94.30} & \textbf{92.51} & \textbf{92.85} & \textbf{70.93} & \textbf{92.61} & \textbf{75.27}\\
    \hline
\end{tabular}
}
\end{table}
In this section we show the benefits of self-distillation for person and vehicle re-id. We indicate the teacher with the name of the backbone (e.g. \tea{50}) and append \quotationmarks{VKD} for its student (e.g. \stf{50}). To validate our ideas, we do not limit the analysis on \resnet{*}; contrarily, we test self-distillation on \teadense{}~\cite{huang2017densely} and \teamobile{} 1.0X~\cite{sandler2018mobilenetv2}. Since learning what and where to look represents an appealing property when dealing with Re-ID tasks~\cite{fu2019STA}, we additionally conduct experiments on \resnet{50} coupled with Bottleneck Attention Modules~\cite{park2018bam} (\resnet{50bam}). 
\\
\noindent Table~\ref{tab:self_distill} reports the comparisons for different backbones: in the vast majority of the settings, \textit{the student outperforms its teacher}. Such a finding is particularly evident when looking at the I2V setting, where the mAP metric gains $4.04\%$ on average. The same holds for the I2I setting on \veri{}, and in part also on V2V. We draw the following remarks: \textit{i)} in accordance with the objective the student seeks to optimise, our proposal leads to greater improvements when few images are available; \textit{ii)} bridging the gap between I2V and V2V does not imply a significant information loss when more frames are available; on the contrary it sometimes results in superior performance; \textit{iii)} the previous considerations hold true across different architectures. As an additional proof, plots from Figure~\ref{fig:plot_ngallery} draw a comparison between models before and after distillation. VKD improves metrics considerably on all three dataset, as highlighted by the bias between the teachers and their corresponding students. Surprisingly, this often applies when comparing lighter students with deeper teachers: as an example, \stf{34} scores better than even \tea{101} on \veri{}, regardless of the number of images sampled for a gallery tracklet.
\subsection{Comparison with State-Of-The-Art}
\newcommand{\marsvTOv}{
\caption{\mars{} \textbf{V2V}}
      \resizebox{\textwidth}{!}{
        \begin{tabular}{llll}
            \hline\noalign{\smallskip}
             Method & \topk{1} & \topk{5} & mAP \\
            \noalign{\smallskip}
            \hline
            \noalign{\smallskip}
            DuATN\cite{si2018DuATM} & 81.2 & 92.5  & 67.7 \\
            TKP\cite{gu2019TKP} & 84.0 & 93.7  & 73.3 \\
            CSACSE+OF\cite{chen2018CSACSE} & 86.3 & 94.7  & 76.1 \\
            STA\cite{fu2019STA} & 86.3 & 95.7  & 80.8 \\
            STE-NVAN\cite{liu2019NVAN} & 88.9 & - & 81.2 \\
            NVAN\cite{liu2019NVAN} & \textbf{90.0} & - & 82.8 \\ 
             \hline
            \stf{50} & 88.7 & 96.1 & 82.2 \\
            \stf{50bam} & 89.4 & \textbf{96.8} & \textbf{83.1} \\
            \hline
    \end{tabular}
    \label{table:SOTAmarsvTOv}
}}
\newcommand{\marsiTOv}{
\caption{\mars{} \textbf{I2V}}
        \resizebox{\textwidth}{!}{
        \begin{tabular}{llll}
            \hline\noalign{\smallskip}
             Method & \topk{1} & \topk{5} & mAP \\
            \noalign{\smallskip}
            \hline
            \noalign{\smallskip}
            P2SNet\cite{wang2017p2snet} & 55.3 & 72.9 & - \\
            Zhang\cite{zhang2017lstm} & 56.5 & 70.6 & - \\
            XQDA\cite{liao2015XQDA} & 67.2 & 81.9 & 54.9 \\
            TKP\cite{gu2019TKP} & 75.6 & 87.6 & 65.1 \\
            STE-NVAN\cite{liu2019NVAN} & 80.3 & - & 68.8 \\ 
            NVAN\cite{liu2019NVAN} & 80.1 & - & 70.2 \\ 
            MGAT\cite{bao2019maskedGAT} & 81.1 & 92.2 & 71.8 \\ 
             \hline
             \stf{50} & 83.9 & 93.2 & 77.3 \\
             \stf{50bam} & \textbf{84.3} & \textbf{93.5} & \textbf{78.1} \\
             \hline
        \end{tabular}
        \label{table:SOTAmarsiTOv}
}}
\newcommand{\dukevTOv}{
\caption{\duke{} \textbf{V2V}}
        \resizebox{\textwidth}{!}{
        \begin{tabular}{llll}
            \hline\noalign{\smallskip}
             Method & \topk{1} & \topk{5}  & mAP \\
            \noalign{\smallskip}
            \hline
            \noalign{\smallskip}
            DuATN\cite{si2018DuATM} & 81.2 & 92.5 & 67.7 \\
            Matiyali\cite{matiyali2019clip-simil} & 89.3 & 98.3 & 88.5 \\
            TKP\cite{gu2019TKP} & 94.0 & - & 91.7 \\
            STE-NVAN\cite{liu2019NVAN} & 95.2 & - & 93.5 \\ 
            STA\cite{fu2019STA} & 96.2 & \textbf{99.3} & 94.9 \\
            NVAN\cite{liu2019NVAN} & \textbf{96.3} & - & \textbf{94.9} \\ 
             \hline
            \stf{50} & 95.0 & 98.9 & 93.4 \\
            \stf{50bam} & 95.2 & 98.6 & 93.5 \\
             \hline
    \end{tabular}
    \label{table:SOTAdukevTOv}
}}
\newcommand{\veriiTOv}{\caption{\veri{} \textbf{I2V}}
\resizebox{\textwidth}{!}{
        \begin{tabular}{llll}
            \hline\noalign{\smallskip}
             Method & \topk{1} & \topk{5}  & mAP \\
            \noalign{\smallskip}
            \hline
            \noalign{\smallskip}
            PROVID\cite{liu2017provid} & 76.8 & 91.4 & 48.5 \\
            VFL-LSTM\cite{alfasly2019VFL-LSTM} & 88.0 & 94.6 & 59.2 \\
            RAM\cite{liu2018ram} & 88.6 & - &  61.5 \\
            VANet\cite{chu2019vanet} & 89.8 & 96.0 &  66.3 \\
            PAMTRI\cite{tang2019pamtri} & 92.9 & 92.9 & 71.9 \\
            SAN\cite{qian2019san} & 93.3 & 97.1 & 72.5 \\
            PROVID-BOT\cite{liu2017provid} & \textbf{96.1} & 97.9 & 77.2 \\
            \hline
            \noalign{\smallskip}
            \stf{50} & 95.2 & \textbf{98.0} & \textbf{82.2} \\
            \stf{50bam} & 95.7 & 98.0 & 81.6 \\
            \hline
        \end{tabular}
        \label{table:SOTAveriiTOv}
    }}
\newcommand{\atrwiTOi}{\caption{\atrw{} \textbf{I2I}}
        \resizebox{\textwidth}{!}{
        \begin{tabular}{llll}
            \hline\noalign{\smallskip}
             Method & \topk{1} & \topk{5} & mAP \\
            \noalign{\smallskip}
            \hline
            \noalign{\smallskip}
            PPbM-a~\cite{li2019atrw} & 82.5 & 93.7 & 62.9 \\
            PPbM-b~\cite{li2019atrw} & 83.3 & 93.2 & 60.3 \\
            NWPU~\cite{yu2019tiger-top-3} & 94.7 & 96.7 & 75.1 \\
            BRL~\cite{liu2019tiger-top2} & 94.0 & 96.7 & 77.0 \\
            NBU~\cite{liu2019tiger-top1} & \textbf{95.6} & \textbf{97.9} & \textbf{81.6} \\
            \hline
            \tea{101} & 92.3 & 93.5 & 75.7 \\
            \stf{101} & 92.0 & 96.4 & 77.2 \\
            \hline
        \end{tabular}
        \label{table:SOTAatrw}
        }}
\newcommand{\dukeiTOv}{
\caption{\duke{} \textbf{I2V}}
        \resizebox{\textwidth}{!}{
        \begin{tabular}{llll}
            \hline\noalign{\smallskip}
             Method & \topk{1} & \topk{5} & mAP \\
            \noalign{\smallskip}
            \hline
            \noalign{\smallskip}
            STE-NVAN\cite{liu2019NVAN} & 42.2 & - & 41.3 \\ 
            TKP\cite{gu2019TKP} & 77.9 & - & 75.9 \\
            NVAN\cite{liu2019NVAN} & 78.4 & - & 76.7 \\ 
            \hline
            \stf{50} & 85.6 & 93.9 & 83.8 \\
            \stf{50bam} & \textbf{86.2} & \textbf{94.2} & \textbf{84.5} \\
            \hline
        \end{tabular}
        \label{table:SOTAdukeiTov}
}}
\begin{table}[t]
\centering
    \begin{minipage}[t]{.32\linewidth}
    \marsiTOv
    \end{minipage}%
    \begin{minipage}[t]{.32\linewidth}
    \dukeiTOv
    \end{minipage} 
    \begin{minipage}[t]{.32\linewidth}
    \veriiTOv
    \end{minipage}
\end{table}
\nicepar{Image-To-Video} Tables~\ref{table:SOTAmarsiTOv}, \ref{table:SOTAdukeiTov} and \ref{table:SOTAveriiTOv} report a thorough comparison with current state-of-the-art (SOTA) methods, on \mars{}, \duke{} and \veri{} respectively. As common practice~\cite{gu2019TKP,bao2019maskedGAT,qian2019san}, we focus our analysis on \resnet{50}, and in particular on its distilled variants \stf{50} and \stf{50bam}. Our method clearly outperforms other competitors, with an increase in mAP w.r.t. top-scorers of 6.3\% on \mars{}, 8.6\% on \duke{} and 5\% on \veri{}. This results is totally in line with our goal of conferring robustness when just a single image is provided as query. In doing so, we do not make any task-specific assumption, thus rendering our proposal easily applicable to both person and vehicle \reid{}.   
\nicepar{Video-To-Video} Analogously, we conduct experiments on the V2V setting and report results in Table~\ref{table:SOTAmarsvTOv} (\mars{}) and Table~\ref{table:SOTAdukevTOv} (\duke{})\footnote{Since \veri{} does not include any tracklet information in the query set, following all other competitors we limit experiments to the I2V setting only.}. Here, VKD yields the following results: on the one hand, on \mars{} it pushes a baseline architecture as \stf{50} close to NVAN and STE-NVAN~\cite{liu2019NVAN}, the latter being tailored for the V2V setting. Moreover -- when exploiting spatial attention modules (\stf{50bam}) -- it establishes new SOTA results, suggesting that a positive transfer occurs when matching tracklets also. On the other hand, the same does not hold true for \duke{}, where exploiting video features as in STA~\cite{fu2019STA} and NVAN appears rewarding. We leave the investigation of further improvements on V2V to future works. As of today, our proposals is the only one guaranteeing consistent and stable results under both I2V and V2V settings.    
\begin{table}[t]
\centering
    \begin{minipage}[t]{0.32\linewidth}
      \marsvTOv
    \end{minipage} 
    \begin{minipage}[t]{0.32\linewidth}
       \dukevTOv
    \end{minipage}%
    \begin{minipage}[t]{0.32\linewidth}
       \atrwiTOi
    \end{minipage} 
\end{table}
\subsection{Analysis on VKD}
\label{sub:vkdablations}
\nicepar{In the absence of camera information.} Here, we address the setting where we do not have access to camera information. As an example, when dealing with animal re-id this information often lacks and datasets come with images and labels solely: can VKD still provide any improvement? We think so, as one can still exploit the visual diversity lying in a bag of randomly sampled images. To demonstrate our claim, we test our proposal on Amur Tigers re-identification (\atrw{}), which was conceived as an Image-To-Image dataset. During comparisons: \textit{i)} since other works do not conform to a unique backbone, here we opt for \tea{101}; \textit{ii)} as common practice in this benchmark~\cite{liu2019tiger-top1,liu2019tiger-top2,yu2019tiger-top-3}, we leverage re-ranking~\cite{zhong2017re}. Table~\ref{table:SOTAatrw} compares VKD against the top scorers in the \quotationmarks{Computer Vision for Wildlife Conservation 2019} competition. Importantly, the student \stf{101} improves over its teacher (1.5\% on mAP and 2.9\% on \topk{5}) and places second behind~\cite{liu2019tiger-top1}, confirming its effectiveness in a challenging scenario. Moreover, we remark that the top-scorer requires additional annotations - such as body parts and pose information - which we do not exploit.   
\nicepar{Distilling viewpoints \textit{vs} time.} Figure~\ref{fig:timevsmulti} shows results of distilling knowledge from multiple views against time (\textit{i.e.} multiple frames from a tracklet). On one side, as multiple views hold more \quotationmarks{\textit{visual variety}}, the student builds a more invariant representation for the identity. On the opposite, a student trained with tracklets still considerably outperforms the teacher. This shows that, albeit the visual variety is reduced, our distillation approach still successfully exploits it. 
\nicepar{VKD reduces the camera bias.} As pointed out in~\cite{tian2018eliminating}, the appearance encoded by a CNN is heavily affected by external factors surrounding the target object (\textit{e.g.} different backgrounds, viewpoints, illumination \ldots). In this respect, is our proposal effective for reducing such a bias? To investigate this aspect, we perform a camera classification test on both the teacher (\textit{e.g.} \tea{34}) and the student network (\textit{e.g.} \stf{34}) by fitting a linear classifier on top of their features, with the aim of predicting the camera the picture is taken from. We freeze all backbone layers and train for 300 epochs ($\text{lr} = 10^{-3}$ and halved every 50 epochs). Table~\ref{tab:camera_ablation} reports performance on the gallery set for different teachers and students. To provide a better understanding, we include a baseline that computes predictions by sampling from the cameras prior distribution. As expected: \textit{i)} the teacher outperforms the baseline, suggesting it is in fact biased towards background conditions; \textit{ii)} the student consistently reduces the bias, confirming VKD encourages the student to focus on identities features and drops viewpoint-specific information. Finally, it is noted that time-based distillation does not yield the bias reduction we observe for VKD (see supplementary materials).
\begin{table}[t]
    \begin{minipage}[t][][b]{.53\linewidth}
    \centering
    \captionof{table}{Analysis on camera bias, in terms of viewpoint classification accuracy.}
    \label{tab:camera_ablation}
    \setlength{\tabcolsep}{5pt}
    \resizebox{0.9\textwidth}{!}{
    \begin{tabular}{l|c|c|c}
    \hline
    \multirow{2}{*}{} & \multicolumn{1}{c|}{\mars{}} & \multicolumn{1}{c|}{\duke{}} & \multicolumn{1}{c}{\veri{}} \\
    \hline
    Prior Class. & 0.19 & 0.14 & 0.06 \\
    \hline
    \tea{34} & \textbf{0.61} & \textbf{0.73} & \textbf{0.55} \\
    \stf{34} & 0.40 & 0.67 & 0.51 \\
    \hline
    \tea{101} & \textbf{0.71} & \textbf{0.72} & \textbf{0.73} \\
    \stf{101} & 0.51 & 0.70 & 0.68 \\
    \hline
    \end{tabular}
    }
    \end{minipage} 
    \hfill
     \begin{minipage}[t][][b]{.45\linewidth}
    \centering
    \includegraphics[width=\linewidth]{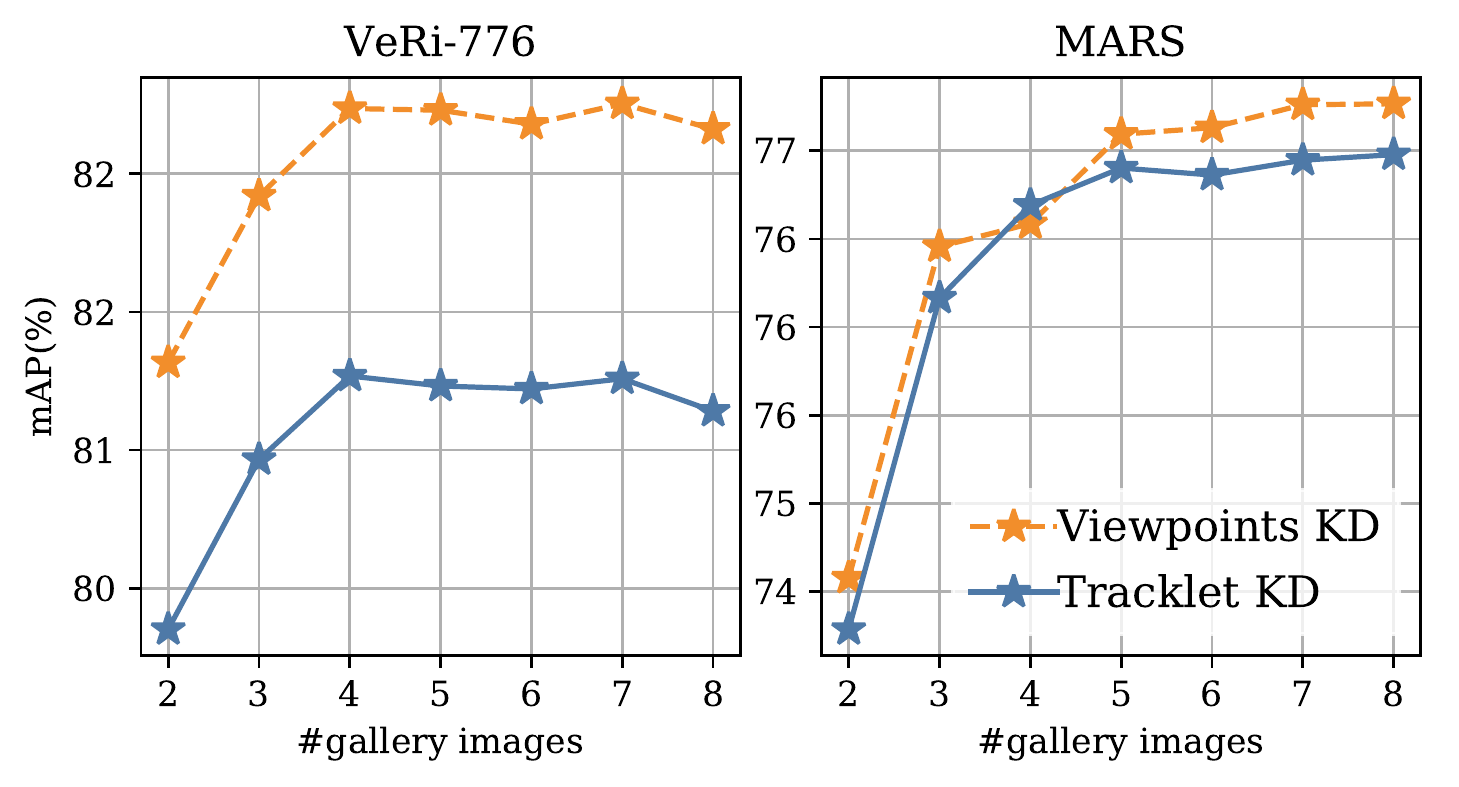}
    \captionof{figure}{Comparison between time and viewpoints distillation.}
    \label{fig:timevsmulti}
    \end{minipage}%
\end{table}
\begin{table}[t]
\caption{Analysis on different modalities for training the teacher.}
\label{tab:teacher_ablation}
\centering
\setlength{\tabcolsep}{2pt}
\resizebox{0.98\textwidth}{!}{
\begin{tabular}{l|c|cc|cc|cc|cc}
    \hline
     & \multirow{3}{*}{Input Bags} & \multicolumn{4}{c|}{\mars{}} & \multicolumn{4}{c}{\duke{}} \\\cline{3-10}
    & & \multicolumn{2}{c|}{I2V} & \multicolumn{2}{c|}{V2V}  & \multicolumn{2}{c|}{I2V} & \multicolumn{2}{c}{V2V}\\
    & & cmc1 & mAP & cmc1 & mAP & cmc1 & mAP & cmc1 & mAP\\
    \hline
    \tea{50} & Viewpoints ($N=2$) & 80.05 & 71.16 & 84.70 & 76.99 & 77.21 & 75.19 & 89.17 & 87.70\\
    \hline
    \tea{50} & Tracklets ($N=2$) & 82.32 & 73.69 & 87.32 & 79.91 & 81.77 & 80.34 & 93.73 & 92.88\\
    \hline
    \stf{50} & Viewpoints ($N=2$) & \textbf{83.89} & \textbf{77.27} & \textbf{88.74} & \textbf{82.22} & \textbf{85.61} & \textbf{83.81} & \textbf{95.01} & \textbf{93.41}\\
    \hline
\end{tabular}
}
\end{table}
\nicepar{Can performance of the student be obtained without distillation?} To highlight the advantages of the two-stage procedure above discussed, we here consider a teacher (\resnet{50}) trained straightly using few frames ($N=2$) only. First two rows of Table~\ref{tab:teacher_ablation} show the performance achieved by this baseline (using tracklets and views respectively). Results show that major improvements come from the teacher-student paradigm we devise (third row), instead of simply reducing the number of input images available to the teacher.
\begin{figure}[t]
\centering
\resizebox{0.9\textwidth}{!}{%
\begin{tabular}{p{0.05\textwidth}llllllllll}
\rotatebox{90}{\quad \quad \quad input} & \attnp{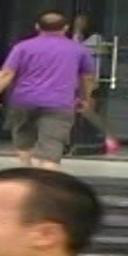} &  \attnp{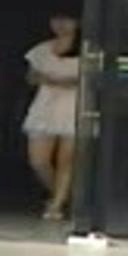} & \attnp{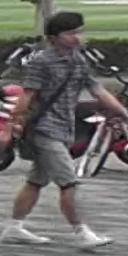} &
\attnp{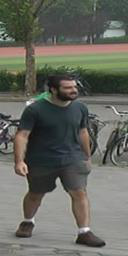} &  \attnp{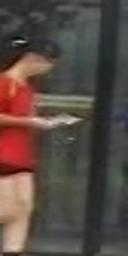} & \attnp{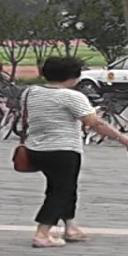} &
\attnp{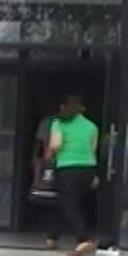} & \attnp{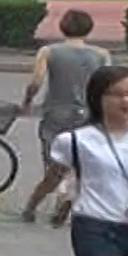} & \attnp{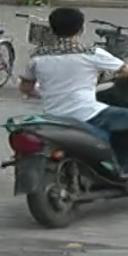} & \attnp{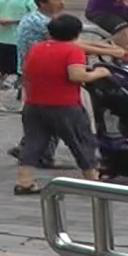} \\\hline \noalign{\vskip 1mm}
\rotatebox{90}{\quad \quad teacher} & \attnp{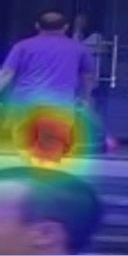} &  \attnp{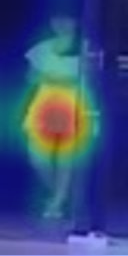} & \attnp{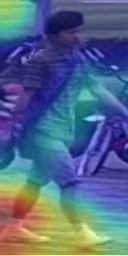} &
\attnp{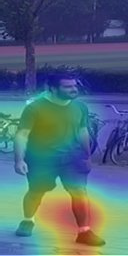} &  \attnp{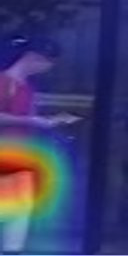} & \attnp{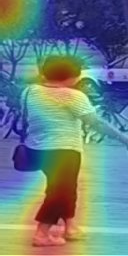} &
\attnp{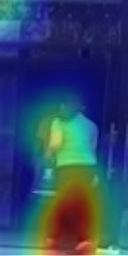} & \attnp{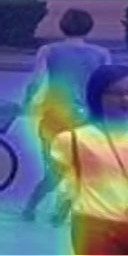} & \attnp{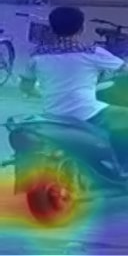} & \attnp{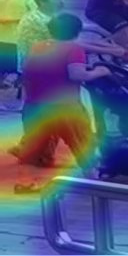} \\\hline \noalign{\vskip 1mm}
\rotatebox{90}{\quad \quad \textbf{student}} & \attnp{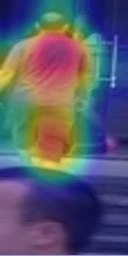} &  \attnp{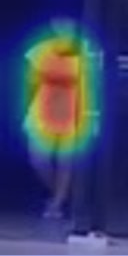} & \attnp{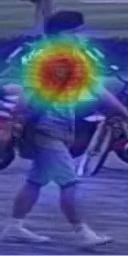} &
\attnp{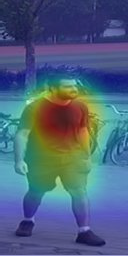} &  \attnp{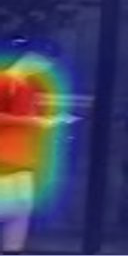} & \attnp{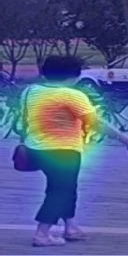} &
\attnp{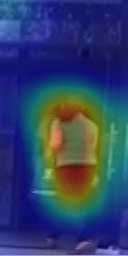} & \attnp{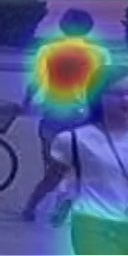} & \attnp{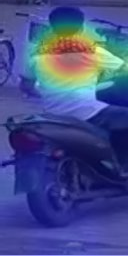} & \attnp{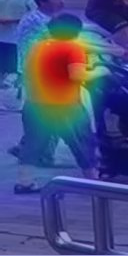} \\
\end{tabular}
}
\vspace{0.02\textwidth}
\resizebox{0.9\textwidth}{!}{%
\begin{tabular}{p{0.05\textwidth}lllllll}
\rotatebox{90}{input} & \attnv{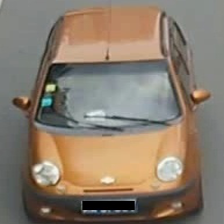} &  \attnv{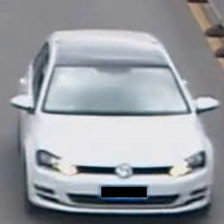} &
\attnv{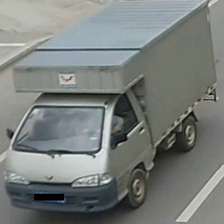} &
\attnv{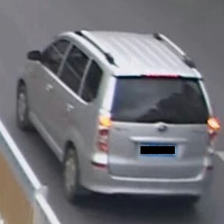} & \attnt{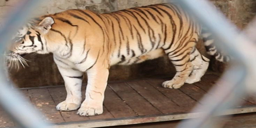} & \attnt{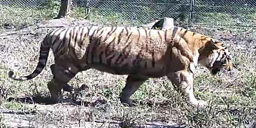} & \attnt{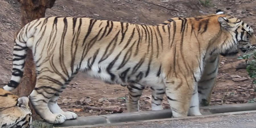}\\
\hline \noalign{\vskip 1mm}
\rotatebox{90}{teacher} & \attnv{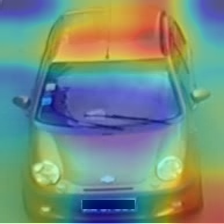} &  \attnv{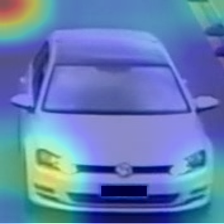} & 
\attnv{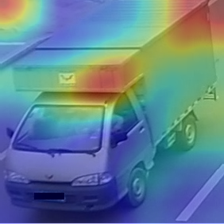} &
\attnv{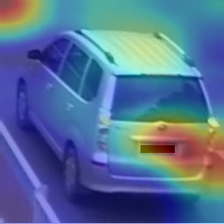} & \attnt{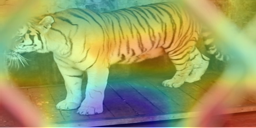} & \attnt{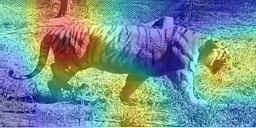} & \attnt{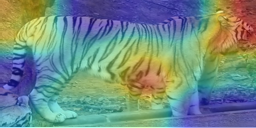}\\
\hline \noalign{\vskip 1mm}
\rotatebox{90}{\textbf{student}} & \attnv{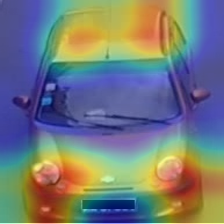} &  \attnv{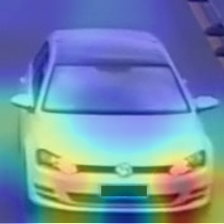} & 
\attnv{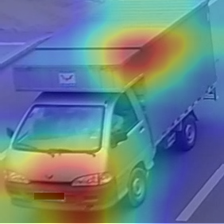} &
\attnv{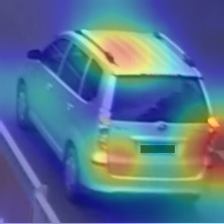} & \attnt{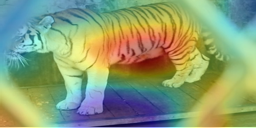} & \attnt{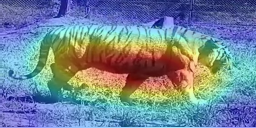} & \attnt{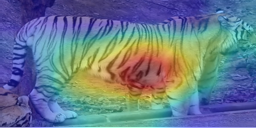}\\
\end{tabular}
}
\caption{Model explanation via GradCam\cite{selvaraju2017gradcam} on \tea{50} (teacher) and \stf{50} (student). The student favours visual details characterising the target, discarding external and uninformative patterns.}
\label{fig:attention}
\end{figure}

\nicepar{Student explanation.} To further assess the differences between teachers and students, we leverage GradCam~\cite{selvaraju2017gradcam} to highlight the input regions that have been considered paramount for predicting the identity. Figure~\ref{fig:attention} depicts the impact of VKD for various examples from \mars{}, \veri{} and \atrw{}. In general, the student network pays more attention to the subject of interest compared to its teacher. For person and animal \reid{}, background features are suppressed (third and last columns) while attention tends to spread to the whole subject (first and fourth columns). When dealing with vehicle \reid{}, one can appreciate how the attention becomes equally distributed on symmetric parts, such as front and rear lights (second, seventh and last columns). Please see supplementary materials for more examples, as well as a qualitative analysis of some of our model errors.
\nicepar{Cross-Distillation.} Differently from other approaches~\cite{bhardwaj2019fewerframes,gu2019TKP}, VKD is not confined to self-distillation, but instead allows the knowledge transfer from a complex architecture (e.g. \resnet{101}) into a simpler one, such as \teamobile{} or \resnet{34} (\textit{cross-distillation}). Here, drawing inspirations from the model compression area, we attempt to reduce the network complexity but, at the same time, increase the profit we already achieve through self-distillation. In this respect, Table~\ref{tab:cross_distill} shows results of cross-distillation, for various combinations of a teacher and a student. It appears that \textit{better the teacher, better the student}: as an example, \stf{34} gains an additional 3\% mAP on \duke{} when educated by \tea{101} rather than \quotationmarks{itself}.
\begin{table}[t]
\caption{Ablation study questioning the impact of each loss term.}
    \centering
    \resizebox{0.98\textwidth}{!}{
    \begin{tabular}{cccccc|cc|cc|cc|cc|cc|cc}
    \hline
    & & \multirow{3}{*}{$\loss{CE}$} & \multirow{3}{*}{$\loss{TR}$} &
    \multirow{3}{*}{$\loss{KL}$} & \multirow{3}{*}{$\loss{DP}$} & \multicolumn{4}{c|}{\mars{}} & \multicolumn{4}{c|}{\duke{}} & \multicolumn{4}{c}{\veri{}}\\\cline{7-18}
    & & & & & & \multicolumn{2}{c|}{I2V} & \multicolumn{2}{c|}{V2V}  & \multicolumn{2}{c|}{I2V} & \multicolumn{2}{c|}{V2V} & \multicolumn{2}{c|}{I2I} & \multicolumn{2}{c}{I2V}\\
    & & & & & & cmc1 & mAP & cmc1 & mAP & cmc1 & mAP & cmc1 & mAP & cmc1 & mAP & cmc1 & mAP\\
    \hline
& & \multicolumn{4}{c|}{\tea{50} (teacher)} & 82.22 & 73.38 & 87.88 & 81.13 & 82.34 & 80.19 & 95.01 & 94.17 & 93.50 & 73.19 & 93.33 & 77.88\\\hline\hline
\multirow{5}{*}{\rotatebox{90}{\stf{50}}} & \multirow{5}{*}{\rotatebox{90}{(students)}} & \cmark & \cmark & \xmark & \xmark & 80.25 & 71.26 & 85.71 & 77.45 & 82.62 & 81.03 & 94.73 & 93.29 & 92.61 & 70.06 & 92.31 & 74.82\\
& & \xmark & \xmark & \cmark & \cmark & 84.09 & \textbf{77.37} & 88.33 & 82.06 & 84.90 & 83.56 & 95.30 & 93.79 & 95.29 & \textbf{79.35} & \textbf{95.29} & \textbf{82.26}\\
& & \cmark & \cmark & \cmark & \xmark & 83.54 & 75.18 & 88.43 & 80.77 & 83.90 & 82.34 & 94.30 & 92.97 & \textbf{95.41} & 78.01 & 95.17 & 81.32\\
& & \cmark & \cmark & \xmark & \cmark & \textbf{84.29} & 76.82 & 88.69 & 81.82 & 85.33 & 83.45 & \textbf{95.44} & \textbf{93.90} & 94.40 & 77.41 & 94.87 & 80.93\\
& & \cmark & \cmark & \cmark & \cmark & 83.89 & 77.27 & \textbf{88.74} & \textbf{82.22} & \textbf{85.61} & \textbf{83.81} & 95.01 & 93.41 & 95.23 & 79.17 & 95.17 & 82.16\\
    \hline
\end{tabular}
}
\label{tab:loss_ablation}
\end{table}
\begin{table}[t]
\caption{Measuring the benefit of VKD for cross-architecture transfer.}
\label{tab:cross_distill}
\centering
\resizebox{0.99\textwidth}{!}{
\setlength{\tabcolsep}{2pt}
\begin{tabular}{ll|cc|cc|cc}
\hline
\multirow{3}{*}{Student {\scriptsize(\#params)}} & \multirow{3}{*}{Teacher {\scriptsize(\#params)}} & \multicolumn{2}{c|}{\mars{}} & \multicolumn{2}{c|}{\duke{}} & \multicolumn{2}{c}{\veri{}}\\\cline{3-8}
& & \multicolumn{2}{c|}{I2V} & \multicolumn{2}{c|}{I2V} & \multicolumn{2}{c}{I2V}\\
& & cmc1 & mAP & cmc1 & mAP & cmc1 & mAP\\
\hline
\multirow{3}{*}{\resnet{34} {\scriptsize(21.2M)}} & \resnet{34} {\scriptsize(21.2M)}& 82.17 & 73.68 & 83.33 & 80.60 & 94.76 & 79.02 \\
 & \resnet{50} {\scriptsize(23.5M)}& 83.08 & 75.45 & 84.05 & 82.61 & \textbf{95.05} & 80.05 \\
 & \resnet{101} {\scriptsize(42.5M)}& \textbf{83.43} & \textbf{75.47} & \textbf{85.75} & \textbf{83.65} & 94.87 & \textbf{80.41} \\
\hline
\hline
\multirow{2}{*}{\resnet{50} {\scriptsize(23.5M)}} & \resnet{50} {\scriptsize(23.5M)}& 83.89 & 77.27 & 85.61 & 83.81 & 95.17 & 82.16 \\
 & \resnet{101} {\scriptsize(42.5M)}& \textbf{84.49} & \textbf{77.47} & \textbf{85.90} & \textbf{84.34} & \textbf{95.41} & \textbf{82.99} \\
\hline 
\hline
\multirow{2}{*}{\teamobile{} {\scriptsize(2.2M)}} & \teamobile{} {\scriptsize(2.2M)}& 83.33 & 73.95 & \textbf{83.76} & 80.83 & 92.61 & 75.27\\
 & \resnet{101} {\scriptsize(42.5M)}& \textbf{83.38} & \textbf{74.72} & \textbf{83.76} & \textbf{81.36} & \textbf{93.03} & \textbf{76.38}\\
 \hline
\end{tabular}
}
\end{table}
\nicepar{On the impact of loss terms.} We perform a thorough ablation study (Table~\ref{tab:loss_ablation}) on the student loss (Eq.~\ref{eq:loss_total}). It is noted that leveraging ground truth solely (second row) hurts performance. Differently, best performance for both metrics are obtained exploiting teacher signal (from the third row onward), with particular emphasis to $\loss{DP}$, which proves to be a fundamental component.
\section{Conclusions}
An effective Re-ID method requires visual descriptors robust to changes in both background appearances and viewpoints. Moreover, its effectiveness should be ensured even for queries composed of a single image. To accomplish these, we proposed Views Knowledge Distillation (VKD), a teacher-student approach where the student observes only a small subset of input views. This strategy encourages the student to discover better representations: as a result, it outperforms its teacher at the end of the training. Importantly, VKD shows robustness on diverse domains (person, vehicle and animal), surpassing by a wide margin the state of the art in I2V. Thanks to extensive analysis, we highlight that the student presents stronger focus on the target and reduces the camera bias.
\nicepar{Acknowledgement} The authors would like to acknowledge Farm4Trade for its financial and technical support.

\newpage
\bibliographystyle{splncs04}
\bibliography{main}
\end{document}


\pagestyle{headings}
\mainmatter
\def\ECCVSubNumber{996}  

\title{Robust Re-Identification by Multiple Views Knowledge Distillation - Supplementary Material} 

\titlerunning{Supplementary Material}

\author{Angelo Porrello \orcidID{0000-0002-9022-8484},
Luca Bergamini \orcidID{0000-0003-1221-8640},
Simone Calderara \orcidID{0000-0001-9056-1538}}
\authorrunning{A. Porrello, L. Bergamini, S. Calderara}
\institute{AImageLab, University of Modena and Reggio Emilia\\
\email{\{angelo.porrello, luca.bergamini24, simone.calderara\}@unimore.it}
}
\maketitle

\begin{table}[b]
    \centering
    \caption{Analysis on camera bias -- in terms of viewpoint classification accuracy -- for different methods. We indicate with \quotationmarks{ResTKD-50} a student restricted to time information solely.}
    \label{tab:biastimevsmv}
    \setlength{\tabcolsep}{5pt}
    \begin{tabular}{l|c|c}
    \hline
    \multirow{2}{*}{} & \multicolumn{1}{c|}{\mars{}} & \multicolumn{1}{c}{\duke{}}\\
    \hline
    Prior Class. & 0.19 & 0.14 \\
    \hline\hline
    \tea{50} (teacher) & \textbf{0.74} & \textbf{0.76}\\
    ResTKD-50 (time-based distillation) & 0.69 & \textbf{0.76}\\
    \stf{50} (viewpoints-based distillation) & 0.49 & 0.69 \\
    \hline
    \end{tabular}
\end{table}
\nicepar{Distilling viewpoints \textit{vs} time: impact on camera bias.} As discussed in the main paper (Introduction and Sec.~4.4), limiting the teacher-student transfer to the temporal axis does not explicitly encourage invariance and robustness to different viewpoints. To further prove such a claim, we again measure the camera bias lying in high-level features, in the same manner as described in Sec.~4.4 of the main paper. This time, though, we focus on a student accessing fewer frames from the same tracklet, thus being educated to capture time information solely. Table~\ref{tab:biastimevsmv} compares this strategy (third row) with our proposal (fourth row), which instead forces the transfer at viewpoint level. As expected: \textit{i)} time-based distillation performs similarly to the teacher, confirming its poor ability to confer robustness to shifts in background appearance; \textit{ii)} as advocated by our work, a student shows a lower camera bias when trained on different viewpoints instead of using temporal information only.
\nicepar{Student explanation - other examples.} In Sec.~4.4 of the main paper, we investigate which regions the student focuses on, showing that it pays higher attention to foreground details when compared to its teacher. We observe that this happens systematically, especially when dealing with person Re-ID. Figure~\ref{fig:attention_suppl} reports additional comparisons between the explanations provided by the teacher and its student on \dukefull{}~\cite{wu2018duke1}.
\begin{figure}[t]
\centering
\resizebox{0.95\textwidth}{!}{%
\begin{tabular}{p{0.05\textwidth}llllllllll}
\rotatebox{90}{\quad \quad \quad input} & \attnpsup{duke_13} &  \attnpsup{duke_3} & \attnpsup{duke_4} &
\attnpsup{duke_5} &  \attnpsup{duke_6} & \attnpsup{duke_7} &
\attnpsup{duke_9} & \attnpsup{duke_11} & \attnpsup{duke_1} & \attnpsup{duke_18} \\\hline \noalign{\vskip 1mm}
\rotatebox{90}{\quad \quad teacher} & \attnpsup{duke_tea_13} &  \attnpsup{duke_tea_3} & \attnpsup{duke_tea_4} &
\attnpsup{duke_tea_5} &  \attnpsup{duke_tea_6} & \attnpsup{duke_tea_7} &
\attnpsup{duke_tea_9} & \attnpsup{duke_tea_11} & \attnpsup{duke_tea_1} & \attnpsup{duke_tea_18} \\\hline \noalign{\vskip 1mm}
\rotatebox{90}{\quad \quad \textbf{student}} & \attnpsup{duke_std_13} &  \attnpsup{duke_std_3} & \attnpsup{duke_std_4} &
\attnpsup{duke_std_5} &  \attnpsup{duke_std_6} & \attnpsup{duke_std_7} &
\attnpsup{duke_std_9} & \attnpsup{duke_std_11} & \attnpsup{duke_std_1} & \attnpsup{duke_std_18} \\
\end{tabular}
}
\caption{Model explanation (\dukefull{}) on \tea{50} (teacher) and \stf{50} (student).}
\label{fig:attention_suppl}
\end{figure}

\begin{figure}[t]
    \begin{minipage}[t][][b]{.65\linewidth}
    \centering
    \includegraphics[width=.99\linewidth]{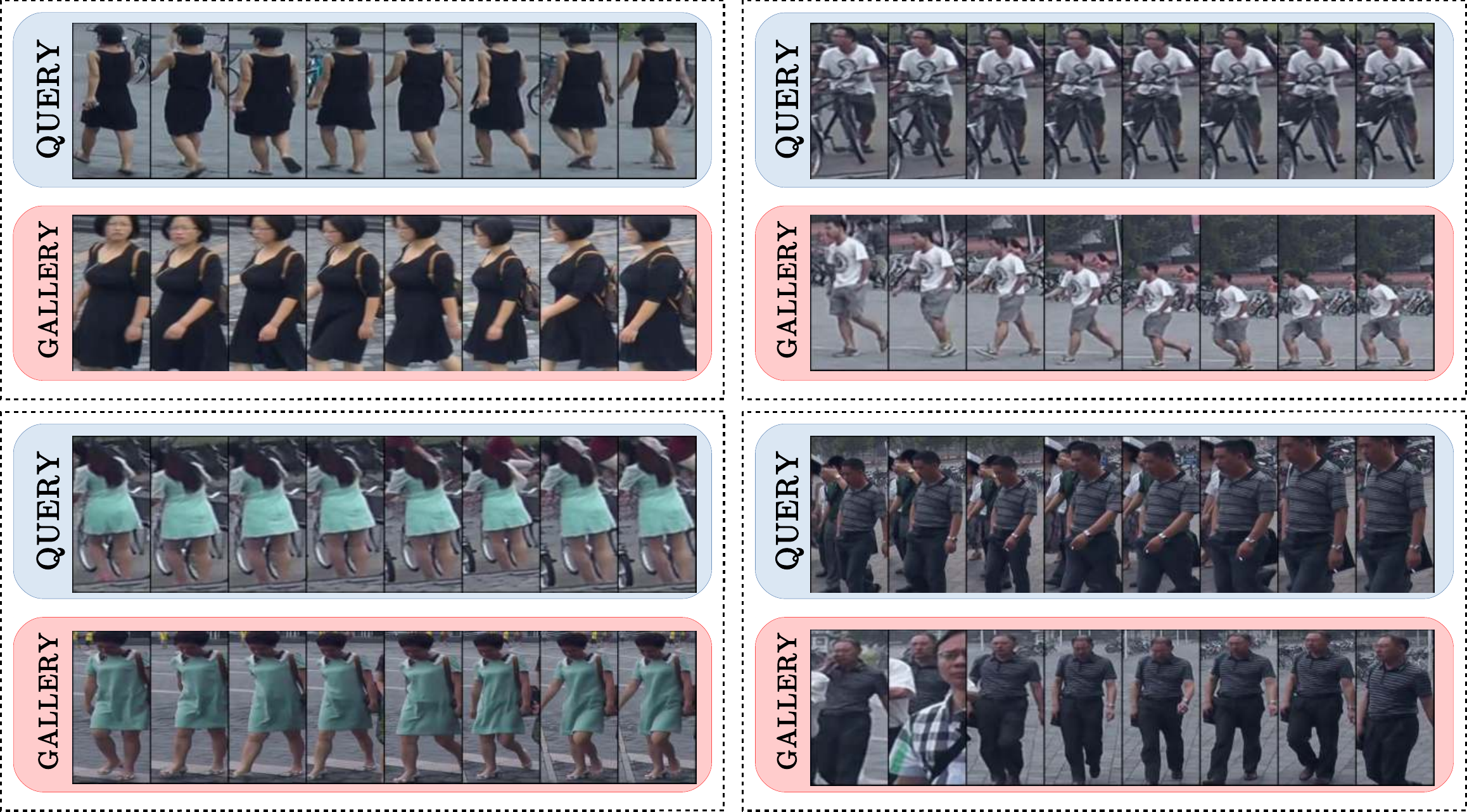}
    \\\small{(a) True Errors}.
    \end{minipage} 
     \begin{minipage}[t][][b]{.34\linewidth}
    \centering
    \includegraphics[width=.99\linewidth]{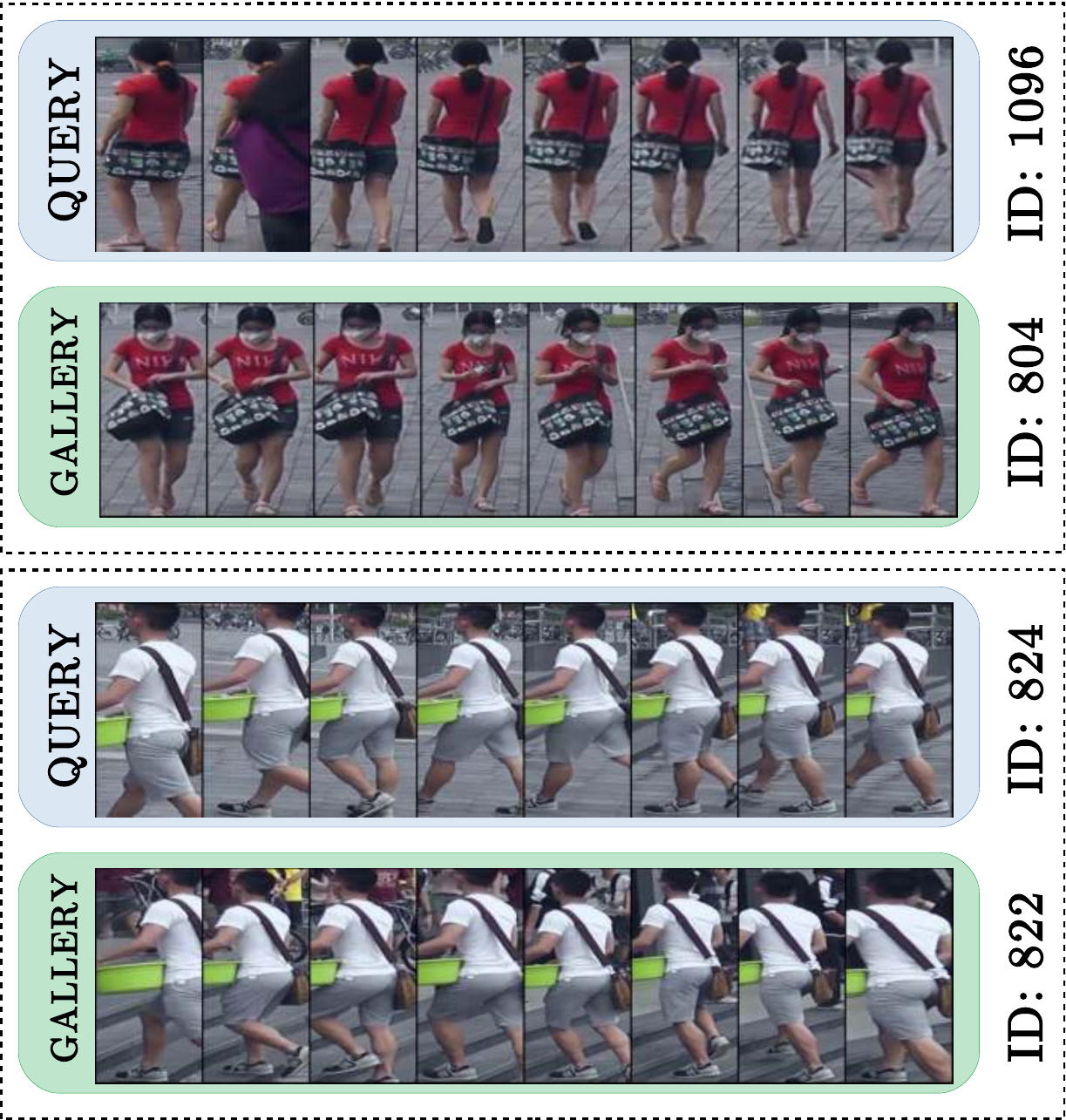}
    \\\small{(b) Wrong ID Annotations}.
    \end{minipage}
    \vskip.5\baselineskip%
    \begin{minipage}[t][][b]{.59\linewidth}
    \centering
    \includegraphics[width=.99\linewidth]{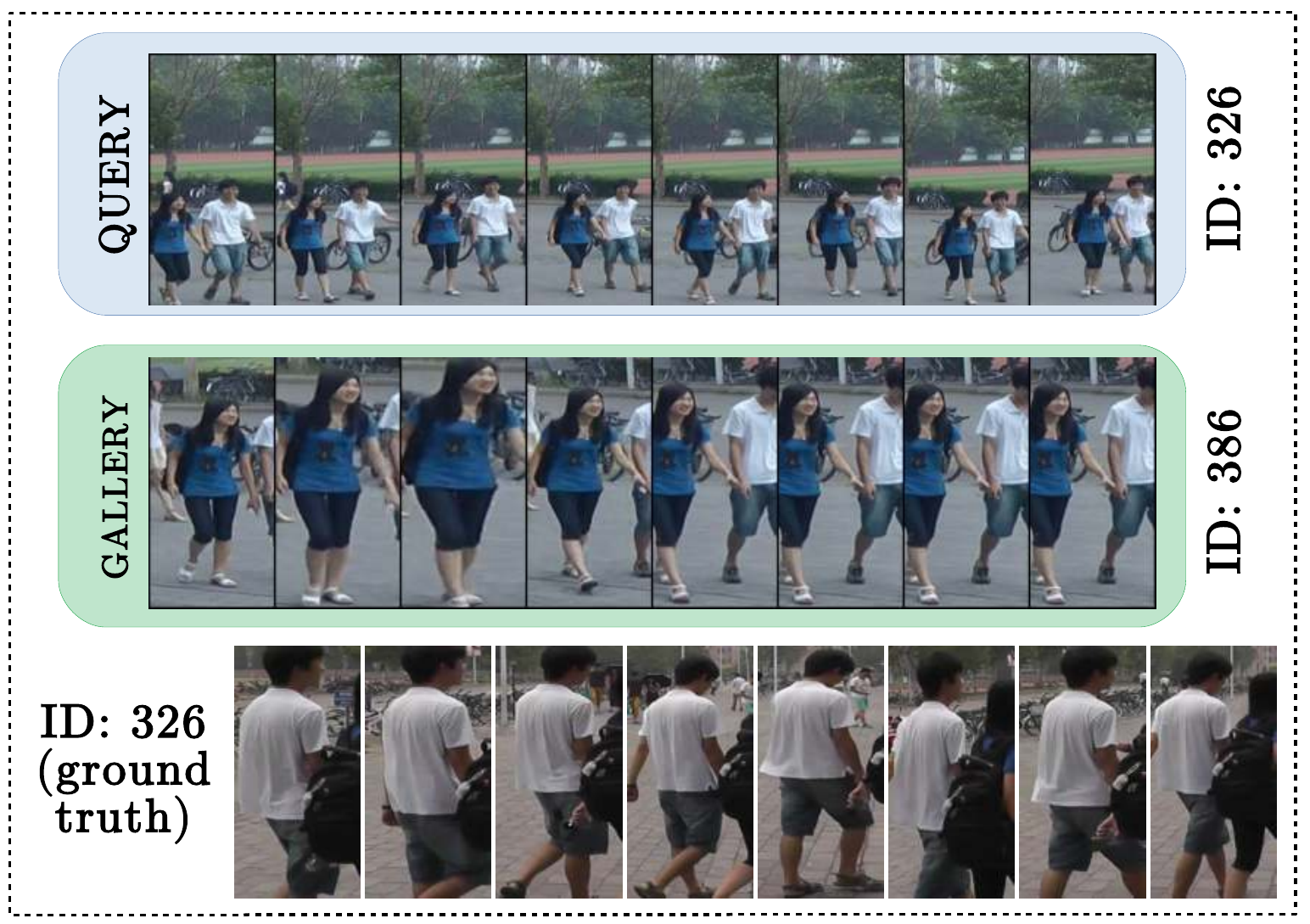} 
    \\\small{(c) Couples of People}.
    \end{minipage} 
     \begin{minipage}[t][][b]{.40\linewidth}
    \centering
    \includegraphics[width=.99\linewidth]{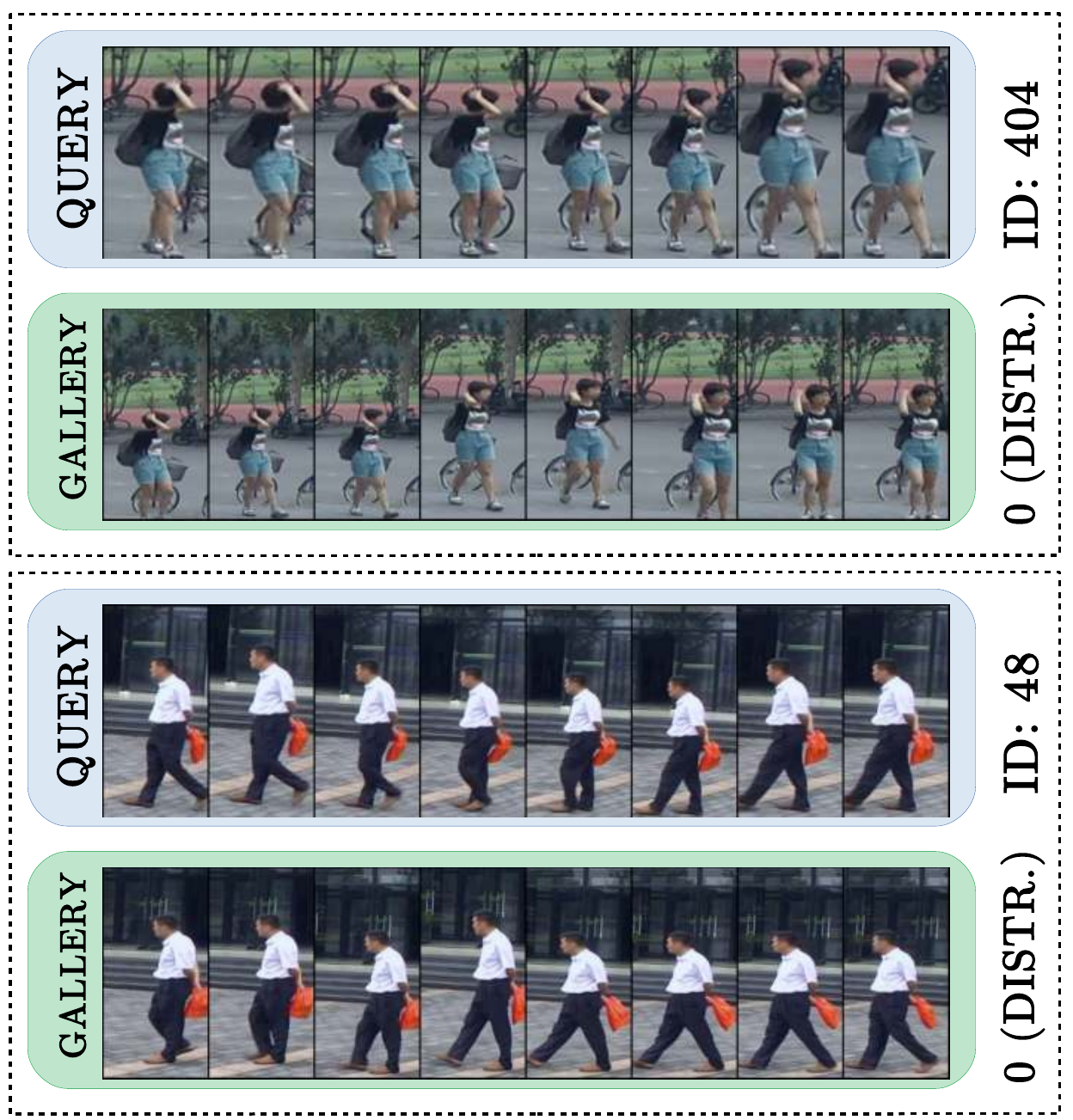} \\\small{(d) Misleading Distractors}.
    \end{minipage}
    \caption{Different categories of errors on \mars{}. While almost half of them can be attributed to our method misidentifying between similar appearances (a), the other half are due to the automatic annotation process. In particular, wrong annotation caused by tracking drift (b), more than one identity in the same tracklet (c) and misleading distractors (d).}
\label{fig:errors}
\hfill
\end{figure}

\nicepar{Errors Analysis} We provide here some visual examples of the errors of our method and try to investigate their nature. With reference to the Video-To-Video setting on \mars{}~\cite{zheng2016mars}, our model (\stf{50}) misidentifies 223 out of 1980 top-1 matchings. From an analysis computed on top of these 223 cases, we identify four different categories of errors. We also asked two external researchers to annotate the errors according to these four classes as follows: 

\begin{enumerate}[a)]
    \item \textbf{True errors}: the network associates the query to a wrong identity from the gallery set (Figure~\ref{fig:errors}a). This often happens when similar clothes and appearances between the two identities fool the network. Out of 223, 103 (\textbf{46.2\%}) were identified as true errors;
    \item \textbf{Wrong ID Annotations}: the ground truth indicates that the network associates the query to a wrong identity from the gallery set. However -- for a limited set of queries -- this does not hold true when visually inspecting the gallery identity. This is due to annotation errors, probably caused by a drift in the tracker (Figure~\ref{fig:errors}b). Out of 223, 29 (\textbf{13.0\%}) were identified as true errors; 
    \item \textbf{Couples of People}: some crops depict more than one subject (\textit{e.g.} two) but only one can be associated with the tracklet id (Figure~\ref{fig:errors}c). Out of 223, 37 (\textbf{16.6\%}) were identified as errors involving frames with more than one person;
    \item \textbf{Misleading Distractors}: cases in which the subject has been correctly identified, but the gallery tracklet was erroneously indicated as a distractor. Again, because this set has not been manually checked, some distractors are valid as they depict people (Figure~\ref{fig:errors}d). Out of 223, 54 (\textbf{24.2\%}) were identified as misleading distractors;
\end{enumerate}
\noindent It is worth noting that the presence of the last three types of errors places a limit on the maximum score a method can obtain.
\bibliographystyle{splncs04}
\bibliography{main}